\documentclass[11pt]{article}
\usepackage[hyperref]{emnlp2018}
\usepackage{times}
\usepackage{url}
\usepackage{latexsym}
\usepackage{ amssymb }
\usepackage{xcolor}
\usepackage{tikz-qtree,tikz-qtree-compat}
\usepackage{diagbox}
\usepackage{amsmath}
\usepackage{sansmath, etoolbox}
\usepackage[textsize=tiny]{todonotes}
\usepackage{float}

\usepackage{textcomp}
\usepackage{footnote}
\usepackage{tablefootnote}

\appto{\sffamily}{\sansmath}
\appto{\rmfamily}{\unsansmath}

\renewcommand{\ss}[1]{\sffamily#1\rmfamily}

\tikzset{every tree node/.style={align=center, anchor=north}}

\aclfinalcopy 

\title{Do Language Models Understand \textit{Anything}?\\{\normalsize On the Ability of LSTMs to Understand Negative Polarity Items}}

\author{Jaap Jumelet \\
  University of Amsterdam \\
  {\tt jaap.jumelet@student.uva.nl} \\\And
  Dieuwke Hupkes \\
  ILLC, University of Amsterdam \\
  {\tt d.hupkes@uva.nl} \\}

\begin{document}
\maketitle

\begin{abstract}
In this paper, we attempt to link the inner workings of a neural language model to linguistic theory, focusing on a complex phenomenon well discussed in formal linguistics: (negative) polarity items.
We briefly discuss the leading hypotheses about the licensing contexts that allow negative polarity items and evaluate to what extent a neural language model has the ability to correctly process a subset of such constructions.
We show that the model finds a relation between the licensing context and the negative polarity item and appears to be aware of the \textit{scope} of this context, which we extract from a parse tree of the sentence.
With this research, we hope to pave the way for other studies linking formal linguistics to deep learning.
\end{abstract}

\section{Introduction}
In the past decade, we have seen a surge in the development of neural language models (LMs). 
As they are more capable of detecting long distance dependencies than traditional n-gram models, they serve as a stronger model for natural language. 
However, it is unclear what kind of properties of language these models encode.
This does not only hinder further progress in the development of new models, but also prevents us from using models as explanatory models and relating them to formal linguistic knowledge of natural language, an aspect we are particularly interested in in the current paper.

Recently, there has been an increasing interest in investigating what kind of linguistic information is represented by neural models, \citep[see, e.g.,][]{conneau2018you,DBLP:journals/corr/LinzenDG16,tran2018importance}, with a strong focus on their \textit{syntactic} abilities.
In particular, \cite{gulordava2018colorless} used the ability of neural LMs to detect noun-verb congruence pairs as a proxy for their awareness of syntactic structure, yielding promising results.
In this paper, we follow up on this research by studying a phenomenon that has received much attention by linguists and for which the model requires -- besides knowledge of syntactic structure -- also a \textit{semantic} understanding of the sentence: negative polarity items (NPIs).

In short, NPIs are a class of words that bear the special feature that they need to be \textit{licensed} by a specific licensing context (LC) (a more elaborate linguistic account of NPIs can be found in the next section).
A common example of an NPI and LC in English are \textit{any} and \textit{not}, respectively: The sentence \textit{He didn't buy any books} is correct, whereas \textit{He did buy any books} is not.
To properly process an NPI construction, a language model must be able to detect a relationship between a licensing context and an NPI.

Following \citet{DBLP:journals/corr/LinzenDG16} and \citet{gulordava2018colorless}, we devise several tasks to assess whether neural LMs (focusing in particular on LSTMs) can handle NPI constructions, and obtain initial positive results.
Additionally, we use diagnostic classifiers \cite{hupkes2018visualisation} to increase our insight in how NPIs are processed by neural LMs, where we look in particular at their understanding of the \textit{scope} of an LCs, an aspect which is also relevant for many other natural language related phenomena.

We obtain positive results focusing on a subset of NPIs that is easily extractable from a parsed corpus but also argue that a more extensive investigation is needed to get a complete view on how NPIs -- whose distribution is highly diverse -- are processed by neural LMs.
With this research and the methods presented in this paper, we hope to pave the way for other studies linking neural language models to linguistic theory. 

In the next section, we will first briefly discuss NPIs from a linguistic perspective.
Then, in Section \ref{sec:data}, we provide the setup of our experiments and describe how we extracted NPI sentences from a parsed corpus. 
In Section \ref{sec:perplexity}, we describe the setup and results of an experiment in which we compare the grammaticality of NPI sentences with and without a licensing context, using the probabilities assigned by the LM.
Our second experiment is outlined in Section \ref{sec:scope}, in which we describe a method for scope detection on the basis of the intermediate sentence embeddings. 
We conclude our findings in Section \ref{sec:discussion}.

\begin{table*}[t]
\begin{center}
\begin{tabular}{|l|l|}\hline
& Context type \\\hline
1. \textit{\underline{Every}} [ \textit{student with \textbf{any} sense} ] \textit{left} & Downward entailing \\
2. \textit{Ann \underline{doubts} that} [ \textit{Bill \textbf{ever} ate any fish} ] & Non-veridical\\
3. \textit{I do\underline{n't}} [ \textit{have \textbf{any} potatoes} ] & Downward entailing\\
4. [ \textit{Did you see \textbf{anybody}} ] \textit{\underline{?}} & Questions
\\\hline
\end{tabular}
\end{center}
\caption{Various example sentences containing NPI constructions. The licensing context scope is denoted by square brackets, the NPI itself in boldface, and the licensing operator is underlined. In our experiments we focus mostly on sentences that are similar to sentence 3.}
\label{npis}
\end{table*}
\section{Negative Polarity Items}\label{sec:npi}

NPIs are a complex yet very common linguistic phenomenon, reported to be found in at least 40 different languages \cite{haspelmath1997indefinite}. 
The complexity of NPIs lies mostly in the highly idiosyncratic nature of the different types of items and licensing contexts. 
Commonly, NPIs occur in contexts that are related to negation and modalities, but they can also appear in imperatives, questions and other types of contexts and sentences. 
This broad range of context types makes it challenging to find a common feature of these contexts, and no overarching theory that describes when NPIs can or cannot occur yet exists \cite{Barker2018}.
In this section, we provide a brief overview of several hypotheses about the different contexts in which NPIs can occur, as well as examples that illustrate that none of these theories are complete in their own regard.
An extensive description of these theories can be found in \citet{giannakidou2008negative}, \citet{hoeksema2008natural}, and \citet{Barker2018}, from which most of the example sentences were taken. 
These sentences are also collected in Table \ref{npis}.

\paragraph{Entailment} 
A downward entailing context is a context that licenses entailment to a subset of the initial clause. 
For example, \textit{Every} is downward entailing, as \textit{Every} [ \textit{student} ] \textit{left} entails that \textit{Every} \mbox{[ \textit{tall student} ]} \textit{left}.
\citet{ladusaw1980polarity} hypothesize that NPIs are licensed by downward entailing contexts. 
Rewriting the previous example to \textit{Every} \mbox{[\textit{ student with any sense} ]} \textit{left} yields a valid expression, contrary to the same sentence with the upward entailing context \textit{some}: \textit{Some} \mbox{[\textit{student with any sense} ]} \textit{left}.
An example of a non-downward entailing context that is a valid NPI licensor is \textit{most}.

\paragraph{Non-veridicality}
A context is non-veridical when the truth value of a proposition (\textit{veridicality}) that occurs inside its scope cannot be inferred. 
An example is the word \textit{doubt}: the sentence \textit{Ann doubts that Bill ate some fish} does not entail \textit{Bill ate some fish}.
\citet{giannakidou1994semantic} hypothesizes that NPIs are licensed only in non-veridical contexts, which correctly predicts that \textit{doubt} is a valid licensing context: \textit{Ann doubts that Bill ate any fish}. 
A counterexample to this hypothesis is the context that is raised by the veridical operator \textit{only}: \textit{Only Bob ate fish} entails \textit{Bob ate fish}, but also licenses \textit{Only Bob ate any fish} \cite{Barker2018}. 

\subsection{Related constructions}
Two grammatical constructions that are closely related to NPIs are Free Choice Items (FCIs) and Positive Polarity Items (PPIs). 

\paragraph{Free Choice Items} FCIs inhibit a property called \textit{freedom of choice} \cite{vendler1967linguistics}, and are licensed in contexts of generic or habitual sentences and modal verbs. An example of such a construction is the generic sentence \textit{Any cat hunts mice}, in which \textit{any} is an FCI. Note that \textit{any} in this case is not licensed by negation, modality, or any of the other licensing contexts for NPIs. English is one of several languages in which a word can be both an FCI and NPI, such as the most common example \textit{any}. Although this research does not focus on FCIs, it is important to note that the somewhat similar distributions of NPIs and FCIs can severely complicate the diagnosis whether we are dealing with an NPI or an FCI. 

\paragraph{Positive Polarity Items}
PPIs are a class of words that are thought to bear the property of scoping above negation \cite{giannakidou2008negative}. Similar to NPIs their contexts are highly idiosyncratic, and the exact nature of their distribution is hard to define. PPIs need to be situated in a veridical (often affirmative) context, and can therefore be considered a counterpart to the class of NPIs. A common example of a PPI is \textit{some}, and the variations thereon. It is shown by \citet{giannakidou2008negative} that there exist multiple interpretations of \textit{some}, influenced by its intonation. The emphatic variant is considered to be a PPI that scopes above negation, while the non-emphatic \textit{some} is interpreted as a regular indefinite article (such as \textit{a}).

\section{Experimental Setup}\label{sec:data}
Our experimental setup consists of 2 phases: first we extract the relevant sentences and NPI constructions from a corpus, and then, after passing the sentences through an LM, we apply several diagnostic tasks to them. 

\subsection{NPI extraction}
For extraction we used the parsed Google Books corpus \cite{googlebooks}. 

We focus on the most common NPI pairs, in which the NPI \textit{any} (or any variation thereon) is licensed by a negative operator (\textit{not}, \textit{n\textquotesingle t}, \textit{never}, or \textit{nobody}), as they can reliably be extracted from a parsed corpus.
As variations of \textit{any} we consider \textit{anybody}, \textit{anyone}, \textit{anymore}, \textit{anything}, \textit{anytime}, and \textit{anywhere} (7 in total including \textit{any}). 

We first identify candidate NPI-LC relations looking only at the surface form of the sentence, by selecting sentences that contain the appropriate lexical items.
We use this as a pre-filtering step for our second method, in which we extract specific subtrees given the parse tree of the sentence. 
We consider 6 different subtrees, that are shown in Table \ref{trees}.

An example of such a subtree that licenses an NPI is the following: 

\begin{center}
\begin{tikzpicture}
\Tree
[.\texttt{VP}
    [.\texttt{VBD} {did} ]
    [.\texttt{RB} {not} ]
    [.\texttt{VP} \edge[roof]; {$\cdots$ \textit{any} $\cdots$} ]
]
\end{tikzpicture}
\end{center}

\noindent which could, for instance, be a subtree of the parse tree of \textit{Bill did not buy any books}. 
In this subtree, the scope of the licensor \textit{not} encompasses the \texttt{VP} of the sentence.
We use this scope to pinpoint the exact range in which an NPI can reside.

\begin{figure}[ht!]
  \centering
  \includegraphics[width=0.95\columnwidth, trim=3mm 3mm 3mm 2mm, clip]{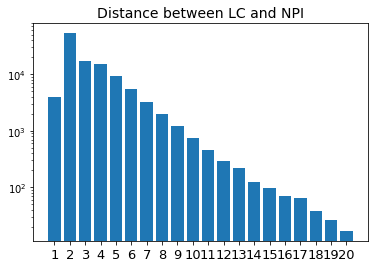}
   \caption{Distribution of distances between NPI and licensing context. Note the log scale on the y-axis.}
  \label{fig:pp_distances}
\end{figure}
Once all NPI constructions have been extracted, we are able to gain more insight in the distance between the licensing operator and an NPI, which we plot in Figure \ref{fig:pp_distances}. 
Note the use of a log scale on the y-axis: in the majority of the constructions (47.2\%) the LC and NPI are situated only 2 positions from each other. 

\begin{table*}[h]
\begin{center}
\begin{tabular}{|l|r l|}\hline
Construction & \# & (\% / corpus)\\
\hline
All corpus sentences & 11.213.916 & \\
Containing any variation of \textit{any} & 301.836 & (2.69\%)\\
Licensed by negative operator & 123.683 & (1.10\%) \\
Detected by subtree extractor & 112.299 & (1.00\%) \\\hline\hline
1. \texttt{(VP (VP \textit{RB} [VP]))} & 70.017 & \\
\multicolumn{1}{|l|}{ \textit{He did \underline{n't}} [ \textit{have \textbf{any} trouble going along} ] \textit{.}}&&\\[3pt]
2. \texttt{(VP (MD \textit{RB} [VP]))} & 27.698 & \\
\multicolumn{1}{|l|}{ \textit{I could \underline{not}} [ \textit{let \textbf{anything} happen to either of them} ] \textit{.}}&&\\[3pt]
3. \texttt{(VP (VP \textit{RB} [NP/PP/ADJP]))} & 8708 & \\
\multicolumn{1}{|l|}{ \textit{"There was \underline{n't}} [ \textit{\textbf{any} doubt in his mind who was preeminent} ] \textit{."}}&&\\[3pt]
4. \texttt{(VP (NP \textit{RB} [VP]))} & 3564 & \\
\multicolumn{1}{|l|}{ \textit{Those words \underline{never}} [ \textit{lead to \textbf{anything} good} ] \textit{.}}&&\\[3pt]
5. \texttt{(S (\textit{RB} [S/SBAR]))} & 1347 & \\
\multicolumn{1}{|l|}{ \textit{The trick is \underline{not}} [ \textit{to process \textbf{any} of the information I encounter} ] \textit{.}}&&\\[3pt]
6. \texttt{(\textit{RB} [NP/PP ADVP])} & 930 & \\
\multicolumn{1}{|l|}{ \textit{There was \underline{not}} [ \textit{a trace of water \textbf{anywhere}} ] \textit{.}}&&\\\hline
\end{tabular}
\end{center}
\caption{Various sentence constructions and their counts that were extracted from the corpus. Similar verb POS tags are grouped under \texttt{VP}, except for modal verbs (\texttt{MD}). LC scope is denoted by square brackets.}
\label{trees}
\end{table*}

\subsection{Model}
For all our experiments, we use a pretrained 2-layer LSTM language model with 650 hidden units made available by \citet{gulordava2018colorless}.\footnote{\url{github.com/facebookresearch/colorlessgreenRNNs/tree/master/data}}
For all tests we used an average hidden final state as initialization, which is computed by passing all sentences in our corpus to the LM, and averaging the hidden states that are returned at the end of each sentence.

We use two different methods to assess the LSTMs ability to handle NPI constructions, which we will discuss in the next two sections: one that is based on the probabilities that are returned by the LM, and one based on its internal activations.

\section{Sentence Grammaticality}\label{sec:perplexity}
In our first series of experiments, we focus on the probabilities that are assigned by the model to different sequences.
More specifically, we compare the exponent of the normalized negative log probability (also referred to as \textit{perplexity}) of different sentences.
The lower the perplexity score of a sentence is, the better a model was able to predict its tokens.

\subsection{Rewriting sentences}
While studying perplexity scores of individual sentences is not very informative, comparing perplexity scores of similar sentences can provide information about which sentence is preferred by the model.
We exploit this by comparing the negative polarity sentences in our corpus with an ungrammatical counterpart, that is created by removing or rewriting the licensing context.\footnote{\textit{Not} and \textit{never} are removed, \textit{nobody} is rewritten to \textit{everybody}.}

To account for the potential effect of rewriting the sentence, we also consider the sentences that originate from replacing the NPI in the original and rewritten sentence with its positive counterpart.
In other words, we replace the variations of \textit{any} by those of \textit{some}: \textit{anything} becomes \textit{something}, \textit{anywhere} becomes \textit{somewhere}, etc.
We refer to these 4 conditions with the terms \ss{\textbf{NPI}$_{neg}$}, \ss{\textbf{NPI}$_{pos}$}, \ss{\textbf{PPI}$_{neg}$} and \ss{\textbf{PPI}$_{pos}$}:

\begin{center}
\begin{tabular}{l l}
\ss{\textbf{NPI}$_{neg}$}:&\textit{Bill did not buy any books}\\
\ss{\textbf{NPI}$_{pos}$}:&* \textit{Bill did buy any books} \\
\ss{\textbf{PPI}$_{neg}$}:&\# \textit{Bill did not buy some books}\\
\ss{\textbf{PPI}$_{pos}$}:&\textit{Bill did buy some books}
\end{tabular}
\end{center}
\ss{\textbf{PPI}$_{neg}$} would be correct when interpreting \textit{some} as indefinite article (\textit{non-emphatic some}). 
In our setup, \ss{\textbf{NPI}$_{neg}$} always refers to the original sentence, as we always use a sentence containing an NPI in a negative context as starting point.
Of the 7 \textit{any} variations, \textit{anymore} is the only one without a PPI counterpart, and these sentences are therefore not considered for this comparison.

\subsection{Comparing sentences}
For all sentences, we compute the perplexity of the original sentence, as well as the perplexity of the 3 rewritten versions of it.
To discard any influence that the removal of the licensing operator might have on its continuation after the occurrence of the NPI, we compute the perplexity of the sentence up to and including the position of the NPI. I.e., in the example of \textit{Bill did not buy any books} the word \textit{books} would not be taken into account when computing the perplexity.

In addition to perplexity, we also consider the conditional probabilities of the PPIs and NPIs, given the preceding sentence.\footnote{We also considered the \textsc{slor} score \cite{pauls2012large}, that was shown in \cite{lau2017grammaticality} to have a strong correlation with human grammaticality judgments. 
The \textsc{slor} score can be seen as a perplexity score that is normalized by the average unigram probability of the sentence.
It turned out, however, that this score had such a strong correlation with the perplexity scores (Spearman's $\rho$ of -0.66, Kendall's $\tau$ of -0.54), that we omitted a further analysis of the outcome. 
}
For example, for \ss{\textbf{NPI}$_{neg}$} we would then compute \textit{P(any  $|$ Bill did not buy)}.

\subsection{Expectations} We posit the following hypotheses about the outcome of the experiments.
\begin{itemize}
\item $PP($\ss{\textbf{NPI}$_{neg}$}$) < PP($\ss{\textbf{NPI}$_{pos}$}$)$:
We expect an NPI construction to have a lower perplexity than the rewritten sentence in which the licensing operator has been removed.
\item $PP($\ss{\textbf{PPI}$_{pos}$}$) < PP($\ss{\textbf{PPI}$_{neg}$}$)$:
Similarly, we expect a PPI to be preferred in the positive counterpart of the sentence, in which no licensing operator occurs.
\item $PP($\ss{\textbf{NPI}$_{neg}$}$) < PP($\ss{\textbf{PPI}$_{neg}$}$)$:
We expect an NPI to be preferred to a PPI inside a negative context.
\item $PP($\ss{\textbf{PPI}$_{pos}$}$) < PP($\ss{\textbf{NPI}$_{pos}$}$)$:
We expect the opposite once the licensor for this context has been removed.
\end{itemize}

\subsection{Results}
\begin{figure}[]
  \centering
  \includegraphics[width=0.95\columnwidth, trim=3mm 2mm 3mm 3mm, clip]{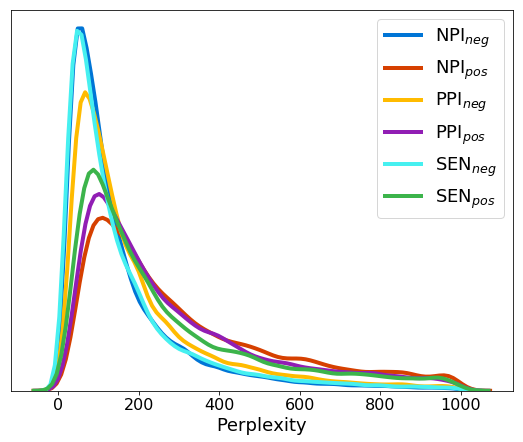}
   \caption{Distribution of perplexity scores for all the sentences.}
  \label{fig:pp_hist}
\end{figure}

In Figure \ref{fig:pp_hist}, we plot the distribution of the perplexity scores for each sentence type.
The perplexities of the original and rewritten sentence without the NPI are indicated by \ss{\textbf{SEN}$_{neg}$} and \ss{\textbf{SEN}$_{pos}$}, respectively. 
This figure shows that the original sentences have the lowest perplexity, whereas the NPIs in a positive context are deemed most improbable by the model.

\begin{figure*}[h!]
\begin{minipage}{0.49\textwidth}
    \centering
\includegraphics[width=\columnwidth]{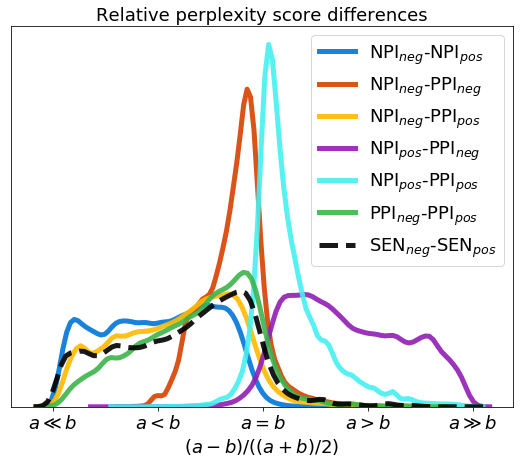}\\
\begin{tabular}{|r|c|c|c|}\hline
\multicolumn{4}{|c|}{\textsc{Perplexity}}\\\hline
$\mathbf{<}$ & \ss{NPI$_{pos}$} & \ss{PPI$_{neg}$} & \ss{PPI$_{pos}$}\\\hline
\ss{NPI$_{neg}$} & 99.2\% & 88.7\% & 95.8\%\\
\ss{NPI$_{pos}$} & - & 3.6\% & 17.3\% \\
\ss{PPI$_{neg}$} & - & - & 91.0\%
\\\hline
\end{tabular}
\end{minipage}
\begin{minipage}{0.49\textwidth}
    \centering
\includegraphics[width=\columnwidth]{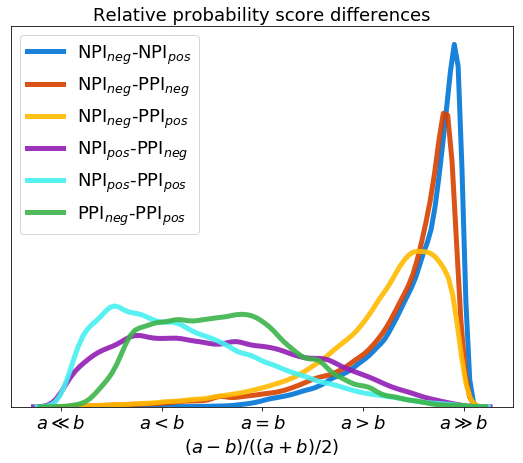}
\begin{tabular}{|r|c|c|c|}\hline
\multicolumn{4}{|c|}{$P(w|c)$}\\\hline
$\mathbf{>}$ & \ss{NPI$_{pos}$} & \ss{PPI$_{neg}$} & \ss{PPI$_{pos}$}\\\hline
\ss{NPI$_{neg}$} & 99.3\% & 94.8\% & 93.4\%\\
\ss{NPI$_{pos}$} & - & 34.0\% & 19.1\% \\
\ss{PPI$_{neg}$} & - & - & 30.1\%
\\\hline
\end{tabular}
\end{minipage}
\caption{Results of perplexity and conditional probability tests. For perplexity a lower score is better, for probability a higher score is better. The plots denote the distribution of the relative differences between the scores of the 6 sentence pairs that are considered.}
\label{perps}
\end{figure*}

More insightful we consider Figure \ref{perps}, in which we plot the distribution of the relative differences of the perplexity scores and conditional probabilities for each of the above mentioned comparisons, and we report the percentage of sentences that complied with our hypotheses.
The relative difference between two values $a$ and $b$, given by $(a-b)/((a+b)/2)$, neatly maps each value pair in a window between -2 $(a\ll b)$ and 2 $(a\gg b)$, thereby providing a better insight in the difference between two arrays of scores.
We highlight some of the previously mentioned comparisons below.

\paragraph{$PP($\ss{NPI$_{neg}$}$) < PP($\ss{NPI$_{pos}$}$)$}
From Figure \ref{perps} it is clear that the model has a very strong preference for NPIs to reside inside the negative scope, an observation that is supported by both the perplexity and probability scores. 
While observable in both plots, this preference is most clearly visible when considering conditional probabilities: the high peak shows that the difference between the probabilities is the most defined of all comparisons that we made.

\paragraph{$PP($\ss{NPI$_{neg}$}$) < PP($\ss{PPI$_{neg}$}$)$}
The model has a strong preference for NPIs over PPIs inside negative scope, although this effect is slightly less prevalent in the perplexity scores. This might be partly due to the fact that there exist interpretations for \textit{some} inside negative scope that are correct (the non-emphatic \textit{some}, as described in Section \ref{sec:npi}). 
When looking solely at the conditional probabilities the preference becomes clearer, showing similar behavior to the difference between \ss{\textbf{NPI}$_{neg}$} and \ss{\textbf{NPI}$_{pos}$}.

\paragraph{$PP($\ss{NPI$_{neg}$}$) < PP($\ss{PPI$_{pos}$}$)$}
The original sentences with NPIs are strongly preferred over the rewritten sentences with PPIs, which indicates that the rewriting in general leads to less probable sentences. 
This finding is confirmed by comparing the perplexities of the original and rewritten sentence \textit{without} the NPI or PPI (dotted line in the left plot in Figure \ref{perps}): the original sentence containing the licensing context has a lower perplexity than the rewritten sentence in 92.7\% of the cases. 
The profile of the differences between the 2 sentences is somewhat similar to the other comparisons in which the negative context is preferred.
Given that the considered sentences were taken from natural data, it is not entirely unsurprising that removing or rewriting a scope operator has a negative impact on the probability of the rest of the sentence.
This observation, however, does urge care when running experiments like this.

\paragraph{$PP($\ss{PPI$_{pos}$}$) < PP($\ss{NPI$_{pos}$}$)$}
When comparing NPIs and PPIs in the rewritten sentences, it turns out that the model does show a clear preference that is not entirely due to a less probable rewriting step. 
Both the perplexity (17.3\%) and probability (19.1\%) show that the NPI did in fact strongly depend on the presence of the licensing operator, and not on other words that it was surrounded with. 
The model is thus able to pick up a signal that makes it prefer a PPI to an NPI in a positive context, even if that positive context was obtained by rewriting it from a negative context.

\paragraph{$PP($\ss{PPI$_{neg}$}$) < PP($\ss{NPI$_{pos}$}$)$}
PPIs in a negative context are strongly preferred to NPIs in a faulty positive context: a lower perplexity was assigned to \ss{\textbf{NPI}$_{pos}$} in only 3.6\% of the cases. 
This shows that the model is less strict on the allowed context for PPIs, which might be related to the non-emphatic variant of \textit{some}, as mentioned before.

\paragraph{$PP($\ss{PPI$_{neg}$}$) < PP($\ss{PPI$_{pos}$}$)$}
A surprising result is the higher perplexity that is assigned to PPIs inside the original negative context compared to PPIs in the rewritten sentence, which is opposite to what we hypothesized. It is especially remarkable considering the fact that the conditional probability indicates an opposite result (at only 30.1\% preference for the original sentence). Once more the outcome of the perplexity comparison might partly be due to the rewriting resulting in a less probable sentence. When solely looking at the conditional probability score, however, we can conclude that the model has a preference for PPIs to reside in positive contexts.

\begin{figure*}[h!]
    \centering
\includegraphics[width=0.8\textwidth]{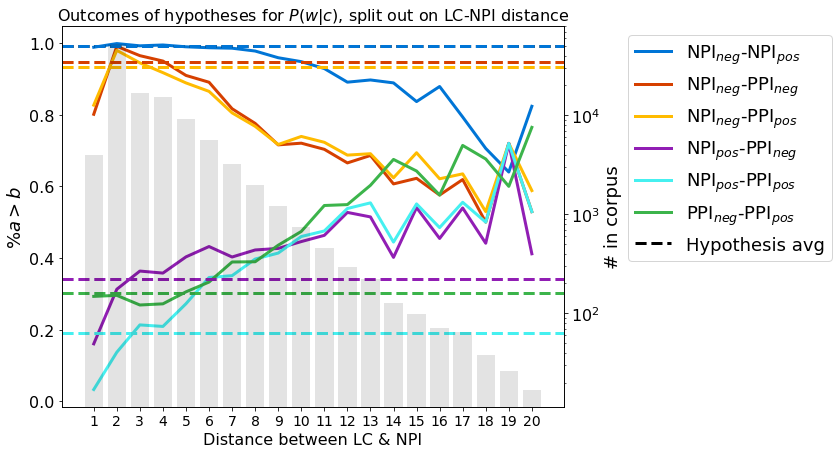}\\
\caption{Outcomes for the conditional probability task, split out on the distance between licensing context and NPI. The averages that are reported in Figure \ref{perps} are denoted by the dotted lines.}
\label{fig:longdist}
\end{figure*}

\paragraph{Long distances}
As shown in Figure \ref{fig:pp_distances}, most distances between the LC and the NPI are rather short. It might therefore be useful to look at the performance of the model on sentences that contain longer distance dependencies. In Figure \ref{fig:longdist} the outcomes of the conditional probability task are split out on the distance between the LC and the NPI.

From this plot it follows that the shorter dependencies were mostly responsible for the outcome of our hypotheses. The significant differences between the original sentence and the rewritten sentences \ss{\textbf{NPI}$_{pos}$} and \ss{\textbf{PPI}$_{neg}$} becomes less defined when the distance is increased.

This might be partly due to the lower occurrence of these constructions: 47.2\% of the sentences in our corpus are situated only 2 positions from each other. Moreover, it would be interesting to see how this behavior matches with that of human judgments.

\paragraph{Conclusion}
We conclude that the LM is able to detect a signal that indicates a strong relationship between an NPI and its licensing context. By comparing the scores between equivalent sentence constructions we were able to account for possible biases of the model, and showed that the output of the model complied with our own hypotheses in almost all cases.

\section{Scope detection} \label{sec:scope}
In the previous section, we assessed the ability of a neural LM to handle NPI constructions, based on the probabilities returned by the LM. 
In the current section, we focus on the hidden states that the LM uses to arrive at a probability distribution over the vocabulary.
In particular, we focus on the \textit{scope} of the licensing operator, which determines where an NPI can occur.

\subsubsection*{Setup}
Using the parse tree extraction method described in Section 3, we annotate all sentences in our corpus with the scope of the licensing operator.
Following \citet{hupkes2018visualisation}, we then train \textit{diagnostic classifiers} to predict for each word in the sentence whether it is inside the licensing scope.
This is done on the basis of the hidden representation of the LM that is obtained after it just processed this word.
We differentiate between 5 different labels: pre-licensing scope words (1), the licensing operator (2), words inside the scope (3), the NPI itself (4), and post-licensing scope words (5). The sentence \textit{The man that died didn't have any relatives, but he died peacefully.}, for example, is annotated as follows:\\

\textit{The$_1$ man$_1$ that$_1$ died$_1$ did$_1$ n't$_2$ have$_3$ any$_4$ relatives$_3$ ,$_5$ but$_5$ he$_5$ died$_5$ peacefully$_5$ .$_5$}\\

The main positions of interest are the transition from within the licensing scope to the post-scope range, and the actual classification of the NPI and LC. 
Of lesser interest are the pre- and post-licensing scope, as these are both diverse embeddings that do not depend directly on the licensing context itself.

We train our model on the intermediate hidden states of the final layer of the LSTM, using a logistic regression classifier. 
The decoder of the LM computes the probability distribution over the vocabulary by a linear projection layer from the final hidden state. By using a linear model for classification (such as logistic regression) we can investigate the expressiveness of the hidden state: if the linear model is able to fulfill a classification task, it could be done by the linear decoding layer too.

As a baseline test, we also train a logistic regression model on representations that were acquired by an additive model using GloVe word embeddings \cite{pennington2014glove}. 
Using these embeddings as a baseline we are able to determine the importance of the language model: if it turns out that the LM does not outperform a simple additive model, this indicates that the LM did not add much syntactic information to the word embeddings themselves (or that no syntactic information is required to solve this task).
We used 300-dimensional word embeddings that were trained on the English Wikipedia corpus (as is our own LM). 

For both tasks (LM and GloVe) we use a subset of 32k NPI sentences which resulted in a total of 250k data points. 
We use a split of 90\% of the data for training, and the other 10\% for testing classification accuracy.

\subsubsection*{Results}
The classifier trained on the hidden states of the LM achieved an accuracy of \textbf{89.7\%} on the test set. The model that was trained on the same dataset using the GloVe baseline scored \textbf{72.5\%}, showing that the information that is encoded by the LM does in fact contribute significantly to this task.
To provide a more qualitative insight into the power of this classifier, we provide 3 remarkable sentences that were classified accurately by the model. Note the correct transition from licensing scope to post-scope, and the correct classification of the NPI and LC in all sentences here.

\begin{enumerate}
\item I$_1$ 'd$_1$ \underline{never$_2$} seen$_3$ \textbf{anything}$_4$ like$_3$ it$_3$ and$_5$ it$_5$ ...$_5$ was$_5$ ...$_5$ beautiful$_5$ .$_5$
\item ``$_1$ I$_1$ do$_1$ \underline{n't$_2$} think$_3$ I$_3$ 'm$_3$ going$_3$ to$_3$ come$_3$ to$_3$ you$_3$ for$_3$ reassurance$_3$ \textbf{anymore}$_4$ ,$_5$ ''$_5$ Sibyl$_5$ grumbled$_5$ .$_5$
\item But$_1$ when$_1$ it$_1$ comes$_1$ to$_1$ you$_1$ ,$_1$ I$_1$ 'm$_1$ \underline{not$_2$} taking$_3$ \textbf{any}$_4$ more$_3$ risks$_3$ than$_3$ we$_3$ have$_3$ to$_3$ .$_5$
\end{enumerate}

We ran a small evaluation on a set of 3000 sentences (47020 tokens), of which 56.8\% were classified completely correctly. Using the GloVe classifier only 22.1\% of the sentences are classified flawlessly.
We describe the classification results in the confusion matrices that are displayed in Figure \ref{conf}.

\begin{figure*}[h]
\begin{tabular}{|r|ccccc|}
\multicolumn{6}{c}{\textit{LSTM Embeddings}}\\\hline
          & \multicolumn{5}{c|}{Correct label}                                                      \\
Pred. & 1              & 2             & 3              & 4             & 5              \\\hline
1         & \textbf{14891} & 83          & 408           & 2             & 760           \\
2         & 203           & \textbf{2870} & 42           & 0             & 59           \\
3         & 850           & 42          & \textbf{14555} & 15          & 1286           \\
4         & 13           & 1          & 32           & \textbf{3005} & 44           \\
5         & 520           & 11          & 821           & 0          & \textbf{6507} \\\hline
Total     & 16477          & 3007          & 15858          & 3022          & 8656         \\\hline
\end{tabular}
\quad
\begin{tabular}{|r|ccccc|}
\multicolumn{6}{c}{\textit{GloVe embeddings}}\\\hline
          & \multicolumn{5}{c|}{Correct label}                                                      \\
Pred. & 1              & 2             & 3              & 4             & 5              \\\hline
1         & \textbf{11166} & 87          & 1077           & 0             & 249           \\
2         & 178           & \textbf{1847} & 82           & 0             & 0           \\
3         & 4708           & 1072          & \textbf{14166} & 353          & 4003           \\
4         & 17           & 0          & 84           & \textbf{2669} & 36           \\
5         & 408           & 1          & 449           & 0          & \textbf{4368} \\\hline
Total     & 16477          & 3007          & 15858          & 3022          & 8656         \\\hline
\end{tabular}
\caption{Confusion matrices for the scope detection task trained on the embeddings of an LSTM and the averages of GloVe embeddings.}
\label{conf}
\end{figure*}

Looking at the results on the LSTM embeddings, it appears that the post-licensing scope tokens (5) were misclassified most frequently: only 75.2\% of those data points were classified correctly. 
The most common misclassification for this class is class 3: an item inside the licensing scope. 
This shows that for some sentences it is hard to distinguish the actual border of the licensing scope, although 90.3\% of the first post-scope embeddings (i.e.\ the first embedding after the scope has ended) were classified correctly. 
The lower performance of the model on this class is mostly due to longer sentences in which a large part of the post-licensing scope was classified incorrectly.
This causes the model to pick up a noisy signal that trips up the predictions for these tokens.
It is promising, however, that the NPIs (4) and licensing operator items (2) themselves are classified with a very high accuracy, as well as the tokens inside the licensing scope (3). 
When comparing this to the performance on the GloVe embeddings, it turns out that that classifier has a strong bias towards the licensing scope class (3). 
This highlights the power of the LSTM embeddings, revealing that is not a trivial task at all to correctly classify the boundaries of the context scope.
We therefore conclude that the information that is relevant to NPI constructions can be accurately extracted from the sentence representations, and furthermore that our neural LM has a significant positive influence on encoding that structural information.

\section{Conclusion}\label{sec:discussion}
We ran several diagnostic tasks to investigate the ability of a neural language model to handle NPIs. 
From the results on the perplexity task we conclude that the model is capable to detect the relationship between an NPI and the licensing contexts that we considered. 
We showed that the language model is able to pick up a distinct signal that indicates a strong relationship between a negative polarity item and its licensing context. 
By comparing the perplexities of the NPI constructions to those of the equivalent PPIs, it follows that removing the licensing operator has a remarkably different effect on the NPIs than on the PPIs.
This effect, however, does seem to vanish when the distance between the NPI and licensing context is increased.
From our scope detection task it followed that the licensing signal that the LM detects can in fact be extracted from the hidden representations, providing further evidence of the ability of the model in handling NPIs. 
There are many other natural language phenomena related to language scope, and we hope that our methods presented here can provide an inspiration for future research, trying to link linguistics theory to neural models.

The setup of our second experiment, for example, would translate easily to the detection of the nuclear scope of quantifiers.
In particular, we believe it would be interesting to look at a wider typological range of NPI constructions, and investigate how our diagnostic tasks translate to other types of such constructions. 
Furthermore, the findings of our experiments could be compared to those of human judgments syntactic gap filling task. These judgments could also provide more insight into the grammaticality of the rewritten sentences.

The hypotheses that are described in Section 2 and several others that are mentioned in the literature on NPIs are strongly based on a specific kind of entailment relation that should hold for the contexts in which NPIs reside.
An interesting follow-up experiment that would provide a stronger link with the literature in formal linguistics on the subject matter, would be based on devising several entailment tasks that are based on the various hypotheses that exists for NPI licensing contexts. 
It would be interesting to see whether the model is able to detect whether a context is downward entailing, for example, or if it has more difficulty identifying non-veridical contexts.
This would then also create a stronger insight in the semantic information that is stored in the encodings of the model.
Such experiments would, however, require the creation of a rich artificial dataset, which would give much more control in determining the inner workings of the LSTM, and is perhaps a necessary step to gain a thorough insight in the LM encodings from a linguistic perspective.

\section*{Acknowledgements}

We thank the reviewers, Samira Abnar, and Willem Zuidema for their useful and constructive feedback.
DH is funded by the Netherlands Organization for Scientific Research (NWO), through a Gravitation Grant 024.001.006 to the Language in Interaction Consortium.

\bibliographystyle{acl_natbib_nourl}
 \bibliography{npi}

\end{document}